\documentclass[conference]{IEEEtran}
\IEEEoverridecommandlockouts
\usepackage{cite}
\usepackage{amsmath,amssymb,amsfonts}
\usepackage{algorithmic}
\usepackage{graphicx}
\usepackage{textcomp}
\usepackage{xcolor}

\usepackage{caption}
\usepackage{amsmath}
\usepackage{lipsum}
\usepackage{mathtools}
\usepackage{cuted}
\usepackage{graphicx}
\usepackage{multirow}
\usepackage{subcaption}
\usepackage{array}

\newcommand\numberthis{\addtocounter{equation}{1}\tag{\theequation}}
\newcommand{\RomanNumeralCaps}[1]
    {\MakeUppercase{\romannumeral #1}}
    
\newcolumntype{P}[1]{>{\centering\arraybackslash}p{#1}}
    
\def\BibTeX{{\rm B\kern-.05em{\sc i\kern-.025em b}\kern-.08em
    T\kern-.1667em\lower.7ex\hbox{E}\kern-.125emX}}
    
\DeclareCaptionType{threshold_calculation}[Heuristic][List of heuristics]

\begin{document}

\bibliographystyle{IEEEtran}

\title{LALR: Theoretical and Experimental validation of Lipschitz Adaptive Learning Rate in Regression and Neural Networks\\
}

\author{
\IEEEauthorblockN{1\textsuperscript{st} Snehanshu Saha}
\IEEEauthorblockA{ \textit{CSIS and APPCAIR} \\
\textit{BITS Pilani K K Birla Goa Campus}\\
Goa, India \\
snehanshu.saha@ieee.org
}
\and
\IEEEauthorblockN{2\textsuperscript{nd} Tejas Prashanth}
\IEEEauthorblockA{\textit{Department of Computer Science} \\
\textit{PES University}\\
Bangalore, India \\
tejuprash@gmail.com
}
\and
\IEEEauthorblockN{3\textsuperscript{rd} Suraj Aralihalli}
\IEEEauthorblockA{\textit{Department of Computer Science} \\
\textit{PES University}\\
Bangalore, India \\
suraj.ara16@gmail.com
}
\and 
\IEEEauthorblockN{4\textsuperscript{th} Sumedh Basarkod}
\IEEEauthorblockA{\textit{Department of Computer Science} \\
\textit{PES University}\\
Bangalore, India \\
sumedhpb8@gmail.com
}
\and 
\IEEEauthorblockN{5\textsuperscript{th} T.S.B. Sudarshan}
\IEEEauthorblockA{\textit{Department of Computer Science} \\
\textit{PES University}\\
Bangalore, India \\
sudarshan@pes.edu
}
\and 
\IEEEauthorblockN{6\textsuperscript{th} Soma S Dhavala}
\IEEEauthorblockA{\textit{Founder} \\
\textit{ML Square}\\
Bangalore, India \\
soma@mlsquare.org
}
}

\maketitle

\begin{abstract}
We propose a theoretical framework for an adaptive learning rate policy for the Mean Absolute Error loss function and Quantile loss function and evaluate its effectiveness for regression tasks. The framework is based on the theory of Lipschitz continuity, specifically utilizing the relationship between learning rate and Lipschitz constant of the loss function. Based on experimentation, we have found that the adaptive learning rate policy enables up to 20x faster convergence compared to a constant learning rate policy. 
\end{abstract}

\begin{IEEEkeywords}
Adaptive learning rate, Lipschitz constant, Mean Absolute Error
\end{IEEEkeywords}

\section{Introduction}
Gradient descent based optimization algorithms, such as Stochastic gradient descent, Adam \cite{kingma2014adam}, have been widely used in the field of deep learning. Gradient descent based learning involves updation of weights of a neural network by back propagation of gradients in order to lower the error\cite{rojas1996backpropagation}. In addition, the training process is a challenging task due to the large number of hyper parameters that require tuning. Among the various hyper-parameters, learning rate is a key factor that influences the speed of convergence of a neural network. Large values of learning rate can hinder convergence and instead may lead to the divergence of the optimization algorithm\cite{bengio2012practical}. Conventionally, learning rate is manually chosen in order to control the rate of convergence. In addition, methods have been developed in order to decay the learning rate over time or use a non-monotonic learning rate scheduler in order for the optimization algorithm to converge faster\cite{seong2018towards}. However, making the learning rate adaptive, is an ongoing research field. 


Neural networks are widely used for a variety of classification and regression based tasks. Regression is implemented in neural networks as a supervised learning problem and is capable of handling complex non-linear dependencies\cite{LEK199639}. A multitude of factors have contributed to the phenomenal success of (deep) neural networks (DNNs). They include, but are not limited to ease of access to big data sets, affordable computing,  plug-and-play deep learning frameworks and auto differentiation frameworks. The Back propagation algorithm and auto differentiation frameworks have greatly simplified the process of fitting DNNs -- it is no longer necessary to write inference software from scratch. Instead, a practitioner or an analyst can simply specify the model, and call the supplied optimization technique, thus abstracting the complexity of the process. While the success of deep neural networks in Computer Vision and Natural Language Processing is transformational, DNNs still suffer from a variety of problems. Some of the criticisms against DNNs are that they can make confident mistakes, due to their opaque nature with respect to the explainability\cite{nguyen2015deep,Rudin}. Making DNNs both reliable and explainable are the new research frontiers. In this paper, we argue that Quantile Regression (QR) can be used to provide prediction intervals, which is one of the ways of quantifying uncertainty\cite{takeuchi2006nonparametric}. In the context of DNNs, QR models can be fit by minimizing the Check Loss function. Moreover, the number of independent regression models that are fit is equivalent to the number of quantiles desired. While one may look for a more efficient architecture to fit all the quantiles simultaneously, fitting multiple independent regression models is a simple alternative. If one can speed-up convergence, then the process of fitting all quantiles independently can be done in nearly the same amount of time that a typical regression model takes to provide point estimates. In this regard, we investigate Adaptive Learning Rates for Check Loss, whose special case is the Mean Absolute Error that estimates conditional medians.

The paper is organized in the following manner. Section \RomanNumeralCaps{2} provides the related work in the field of convergence of gradient descent. Section \RomanNumeralCaps{3} describes the motivation and the contributions of the paper. Section \RomanNumeralCaps{4} provides a theoretical background on Lipschitz continuity and its applications in neural networks. Section \RomanNumeralCaps{5} provides details on the derivation of Lipschitz constant for Mean Absolute Error and Quantile loss. Section \RomanNumeralCaps{6} and \RomanNumeralCaps{7} provide the experimental results for the theoretical framework. Section \RomanNumeralCaps{8} provides a conclusion and future areas of research for this work.  

\section{Related work}
Various approaches have been attempted in order to obtain faster convergence. Many of these approaches involve a theoretical study of gradient descent based optimization algorithms. For example, Hardt et. al. \cite{DBLP:journals/corr/HardtRS15} worked on the theoretical proof for stability of Stochastic Gradient Descent(SGD). The relationship between generalisation error and stability was identified and a stability measure was defined. Similarly, Kuzborskij et. al.\cite{kuzborskij2017data} studied the data dependent stability of stochastic gradient descent. The data-dependent notion of algorithmic stability was established and used to employ generalisation bounds. In addition, generalisation bounds were computed based on data-dependent distribution and initialisation of SGD. 
Development of novel optimization algorithms is another type of solution worthy of mention. For example, Adam \cite{kingma2014adam} optimization algorithm enabled deep learning models to achieve faster convergence. The core idea behind Adam was to combine the usage of momentum-based gradient descent and RMSProp \cite{ruder2016overview}. 

\section{Motivation \& contribution}
Statistical Software such as {\texttt{SAS}},{\texttt{STATA}}, and many libraries in \texttt{R}, provide uncertainty estimates, along with predictions, in terms of standard errors and p-values. Bayesian counterparts produce credible intervals. The classical Frequentist estimation techniques either rely on second-order optimization techniques to produce confidence intervals or rely on special modeling assumptions. The Bayesian analogues require access to efficient, cheap, posterior samples, based on which any functional of the random variables can be computed. The statistics community lays enormous importance on producing such inference summaries. Unfortunately, Machine Learning tools rarely provide such reports, as they are primarily concerned with prediction tasks alone. In the context of DNNs, the problem is more pronounced, as the current Deep Learning landscape is still evolving with respect to providing such uncertainty estimates. For instance, L-BFGS technique is still experimental in \texttt{pytorch}. Developing generic, rich inference techniques is one way to make Deep Learning credible. Another avenue is to consider uncertainty quantification as primary inference goal problem that can work with the almost ubiquitous Back Propagation and Stochastic Gradient Descent. One such method is Quantile Regression(QR). 
QR is well-known in the field of econometrics, and has been introduced to the ML community fairly recently \cite{takeuchi2006nonparametric}. The benefit of QR applied in the context of DNNs is that no new inferential algorithms are required to fit them -- one only needs Check Loss, also called as Quantile loss. It is desirable to look for methods to accelerate the convergence by  exploiting the Check loss function structure, which can support the workhorse inference techniques.

\par We propose an adaptive learning rate scheme for training neural networks with Mean Absolute Error(MAE) and Check loss as the loss functions. Noting that MAE is a special case of Check loss, an adaptive learning rate scheme for Check loss is also proposed. 
Unlike mean squared error(MSE), MAE as a loss function is robust to outliers since it relies on the absolute value of errors instead of the square of error\cite{reich2016case}. In addition, since MAE is primarily applied for regression tasks, the adaptive learning rate is derived for regression based problems in neural networks.  
We contribute to the following problems
\begin{enumerate}
    \item Theoretical framework for computation of Lipschitz adaptive learning rate (LALR) for MAE loss function in single and multi label multivariate regression models, including Quantile regression
    \item Compute LALR for Check loss function using neural networks
    \item Evaluate the effectiveness of the framework against regression based data sets
\end{enumerate}{}

\section{Theoretical background}

\subsection{Notation}
We use the following notation
\begin{enumerate}
    \item m indicates the batch size and n indicates the number of predicted output values in multivariate regression. 
    \item ($x^i,y^i$) refers to a single training example. ($x^i_j,y^i_j$) refers to a particular output value for a single training example.
    \item A superscript of $l$, such as $a^{[l]}$, denotes the layer number and a superscript of L denotes the last layer, unless specified otherwise. For instance, $y^{[L]}$ denotes the actual output of the last layer according to the training data. 
    \item $a^{[l]}$ is used to represent the activation at a particular layer, with a subscript indicating the activation of a particular neuron in that layer. For example, $a^{[l]}_j$ represents the activation value of the $j^{th}$ neuron in layer $l$. The same notation is followed for $y^{[l]}$, which indicates the correct output at a particular layer. 
    \item $W^{[l]}_{ij}$ denotes the weights from neuron $i$ in layer $l-1$ and neuron $j$ in layer $l$.
\end{enumerate}{}

\subsection{Lipschitz continuity}
A function $f(x)$ is said to be Lipschitz continuous in its domain if there exists a constant $k$ in its domain such that for every pair of points, the absolute value of the slope between those points is not greater than $k$. The minimum value of $k$ is known as the Lipschitz constant. Mathematically, the Lipschitz constant for a function $f$,that depends on $x$,is expressed as $\left\Vert f(x_1) - f(x_2)\right\Vert \leq k \left\Vert x_1 - x_2 \right\Vert$
where $k$ is the Lipschitz constant. Since MAE is Lipschitz continuous in its domain, there exists a Lipschitz constant $k$ in its domain. Since the mean value theorem holds good, the supremum of the gradient, sup  $  \left\Vert \nabla f(x) \right\Vert$, exists and the supremum of the gradient is one such Lipschitz constant. 

By computing the Lipschitz constant of the loss function, $ max \left\Vert \nabla_{w} f \right\Vert$, one can constrain  the change in weights in the weight update rule to $\triangle w \leq 1$ by setting the learning rate to be equal to the reciprocal of the Lipschitz constant.

\begin{center}
    $\textbf{w}$ = $\textbf{w}$ $ - $  $\eta . \nabla_{w} f $
\end{center}{}
where $\eta = \frac{1}{max \left\Vert \nabla_{w} f \right\Vert} $. This particular choice of LALR, under the assumption that gradients cannot change arbitrarily fast, ensures a convex quadratic upper bound, minimized by the descent step. It is fairly straightforward to show (via Taylor series expansion of $f$), $f\in C^2$ that ${f(w^{k+1})} \leq {f(w^{k})} - \frac{1}{2L} \left\Vert \nabla{f(w^{k})} \right\Vert^{2} $. This implies that Gradient descent decreases $f$ if $\eta=1/L$ where $L$ is the Lipschitz constant.


\subsection{Lipschitz constant in neural network}
In a neural network, the gradients are smaller in the earlier layers than in the last layer. Consequently, the following relation holds true \cite{DBLP:journals/corr/HardtRS15},

\begin{center}
    $\max_{ij} \left\Vert \frac{\partial E}{\partial w_{ij}^{[L]}} \right\Vert$ $\geq$  $ \left\Vert \frac{\partial E}{\partial w_{ij}^{[l]}} \right\Vert$ $\forall l,i,j$
\end{center}

Hence, the maximum value of the gradient in the neural network can be found using the maximum value of gradient in the last layer.

\subsection{Quantile regression}
The most common loss function used in regression is the Mean Squared Error(MSE). It can be shown that, minimizing MSE is equivalent is maximizing the log-likelihood under the Gaussian noise assumption i.e.  $y_i = f(x_i) + \epsilon_i;
    \epsilon_i = N(0,\sigma^2)$.
Consequently, we get $E[y|x] = f(x)$, where $E[.]$ is the expectation operator. It means that, the minimization of MSE leads to the conditional mean. However, it is known that mean, as a measure of location, is not robust to outliers, due to which median is preferred in such cases. It is also known that, median is the minimization of MAE. In addition, generalization of the MAE is the Check loss, which is given by: $L_{\tau}(e) = (\tau - I(e<0))e$.
Similar to the way in which MSE is shown to maximize the likelihood under Gaussian noise assumption, Check loss can be shown to maximize the log-likelihood under Asymmetric Laplace noise(ALD) assumption: $ y_i = f(x_i) + \epsilon;
    \epsilon = ALD(0,1,\tau)$ \& $ALD(y;\mu,\sigma,\tau) \equiv \frac{\tau(1-\tau)}{\sigma} \exp(-\rho_{\tau}(\frac{y-\mu}{\sigma}))$.
Thereafter, it can be shown that $P(y \le \mu) = \tau$ so that the predicted value can be interpreted as the corresponding conditional quantile. By fitting multiple quantiles, a prediction interval of required coverage can be constructed.

\section{Mathematical derivation}

\subsection{Multiple regression using neural networks}
The following section provides the derivation for the Lipschitz constant for the Mean Absolute Error as the loss function. Assuming one output variable, MAE is given by $E(a^{[L]}, y)=\frac{1}{m} \sum\limits_{i=1}^m \lvert a^{(i)[L]} - y^{(i)}\rvert$ where $m$ is the batch size.

Consider a subset of a batch of $m$ training examples $(\textbf{x}^{(i)}, y^{(i)})$, say $m_1$, which represents the training examples for which $a^{(i)[L]} > y^{(i)}$. Similarly, let $m_2$ be the training examples in the batch for which $a^{(i)[L]} < y^{(i)}$. 

$E(a^{[L]}, y)$ = $\frac{1}{m} \sum\limits_{\substack{i=1 \\ (\textbf{x}^{(i)}, y^{(i)}) \in m_1}}^m \left( a^{(i)[L]} - y^{(i)} \right) + \frac{1}{m} \sum\limits_{\substack{i=1 \\ (\textbf{x}^{(i)}, y^{(i)}) \in m_2}}^m \left( y^{(i)} - a^{(i)[L]} \right) $

Let $a^{[L]}$ and $b^{[L]}$ be two sets of predicted values for output for two different sets of weight matrices of the neural network. Then, $E(a^L, y)-E(b^L,y)=
    \frac{1}{m} \sum\limits_{i=1}^m \lvert a^{(i)[L]} - y^{(i)}\rvert - \lvert b^{(i)[L]} - y^{(i)}\rvert$
. The equation can be elaborated by considering four different cases based on the values of $\boldsymbol{a^L - y}$ and $\boldsymbol{b^L - y}$ . 
Representing the equation in the form of vectors of dimensions $m \times 1$, each case represents a subset of values, with only those components of the vectors activated that match each case. 
We simplify the equation as follows\footnote{Note: Although MAE is not twice differentiable, the functions obtained in each of the four cases are twice differentiable, $f\in C^2$, thus abiding by the assumption behind the proof described in Section \RomanNumeralCaps{4b} }:
\newline
Case 1: $(x_i,y_i)\in(m_1,n_1)$ that satisfy the conditions $\boldsymbol{(a^L-y)_{m_1,n_1}>0}$ and $\boldsymbol{(b^L-y)_{m_1,n_1}>0}$
\begin{align*}
 \frac{1}{m} ((\boldsymbol{a^L} - \boldsymbol{y}) - (\boldsymbol{b^L} - \boldsymbol{y}))_{m1,n1}= \frac{1}{m} (\boldsymbol{a^L} - \boldsymbol{b^L})_{m1,n1} \numberthis
\end{align*}\\
Similarly, the equation is simplified for the remaining cases and an inequality expression is established, as done in Case 1. 
Since $\boldsymbol{a^L}$ and $\boldsymbol{b^L}$ are mutually exclusive among the four cases, adding equations for each of the four cases yields the original $\boldsymbol{a^L}$ and $\boldsymbol{b^L}$. Hence,

\begin{equation}
    \boldsymbol{E(a^L)-E(b^L)} \leq \frac{1}{m} \boldsymbol{(a^L - b^L)}
\end{equation}

After applying L1-norm,

\begin{equation}
    \frac{||\boldsymbol{E(a^L)-E(b^l)}||}{||\boldsymbol{(a^L - b^L)}||} \leq \frac{1}{m}
\end{equation}



Considering the following backpropagation equation, 
\begin{equation*}
    \max_{ij} \lvert \frac{\partial E}{\partial w_{ij}^{[L]}} \rvert \leq \max_{ij}| \frac{\partial E}{\partial a_{j}^{[L]}}| . \max_{ij}| \frac{\partial a_j^{[L]}}{\partial z_{j}^{[L]}}| . \max_{ij}| \frac{\partial z_j^{[L]}}{\partial w_{ij}^{[L]}}|
\end{equation*}

\begin{equation*}
    \max_{ij}| \frac{\partial E}{\partial w_{ij}^{[L]}}| \leq \max_{ij}| \frac{\partial E}{\partial a_{j}^{[L]}}| . \max_{ij}| \frac{\partial a_j^{[L]}}{\partial z_{j}^{[L]}}| . \max_{j}|a^{[L-1]}_j|
\end{equation*}

$\max_{ij}| \frac{\partial a_j^{[L]}}{\partial z_{j}^{[L]}}|$ can be considered to be 1  if the final layer activation function is ReLU for the  regression model. Let $K_z$=$\max_{j}|a^{[L-1]}_j|$; then, $\max_{ij}| \frac{\partial E}{\partial w_{ij}^{[L]}}| \leq \frac{K_z}{m}$.
Hence, the Lipschitz constant is equal to
\begin{equation}
    \frac{K_z}{m}
\end{equation}

\subsection{Multi-label Multivariate regression using neural networks}
The following section provides the derivation for the Lipschitz constant for the Mean Absolute Error as the loss function for multivariate regression. The MAE is: $E(a^L, y)=\frac{1}{mn} \sum\limits_{i=1}^m \sum\limits_{j=1}^n \lvert a^{(i)[L]}_{j} - y^{(i)}_{j}\rvert $, where $m$ is the batch size and $n$ is the number of labels.

Consider $\boldsymbol{a^L}$, the predicted output of the neural network for multivariate regression, to be a flattened form of the conventional matrix of size m $\times$  n. Let the output be of size $(m \cdot n) \times 1$. Consider \textbf{y} to be a flattened vector of size $(m \cdot n) \times 1$ . 
Consider a subset of a batch of $m$ training examples $({x}^{(i)}_{j}, y^{(i)}_{j})$, say $mn_1$, which represents the corresponding output units for the training examples for which $a^{(i)[L]} > y^{(i)}$. Similarly, let $mn_2$ be the corresponding output units in the batch for which $a^{(i)[L]} < y^{(i)}$. 
\begin{align*}
    E(a^{[L]}, y) = \frac{1}{mn} \sum\limits_{\substack{i=1 \\ ({x}^{(i)}_{j}, y^{(i)}_{j}) \in mn_1}}^{mn} \left( a^{(i)[L]} - y^{(i)} \right) +\\
\frac{1}{mn} \sum\limits_{\substack{i=1 \\ ({x}^{(i)}_{j}, y^{(i)}_{j}) \in mn_2}}^{mn} \left( y^{(i)} - a^{(i)[L]} \right)
\end{align*}{}

Let $a^{[L]}$ and $b^{[L]}$ be two sets of predicted values for output for two different sets of weight matrices of the neural network. Then, $ E(a^L, y)-E(b^L,y)=\frac{1}{mn} \sum\limits_{i=1}^{mn} \lvert a^{(i)[L]} - y^{(i)}\rvert - \lvert b^{(i)[L]} - y^{(i)} \rvert$. The proof is carried out in the same manner, as described in Section \RomanNumeralCaps{5A}.
The Lipschitz constant is obtained to be
\begin{equation}
    \frac{K_z}{mn}
\end{equation}

\subsection{Check loss using neural networks}
\[
    \rho_{\tau}(x)= 
\begin{cases}
    x \tau & \text{if } x\ge 0\\
    -x (1-\tau),              & \text{otherwise}
\end{cases}
\]

Without loss of generality, assume that $x_1 < x_2$

Case-1: $ 0 < x_1 < x_2$
\begin{eqnarray*}
    && \implies \frac{|\rho_{\tau}(x_2) - \rho_{\tau}(x_1)|}{|x_2 - x_1|} \le \tau  
\end{eqnarray*}

Case-2: $  x_1 < 0 < x_2$
\begin{eqnarray*}
    && \implies \frac{|\rho_{\tau}(x_2) - \rho_{\tau}(x_1)|}{|x_2 - x_1|} \le \tau 
\end{eqnarray*}

Case-3: $  x_1 <  x_2 < 0$
\begin{eqnarray*}
    && \implies \frac{|\rho_{\tau}(x_2) - \rho_{\tau}(x_1)|}{|x_2 - x_1|} \le (1-\tau)
\end{eqnarray*}

$\therefore L_{\rho_{\tau}(.)} = \max(\tau,1-\tau) \because \tau \in [0,1]$

Hence, the Lipschitz constant is
\begin{equation}
    \frac{K_z * \max(\tau,1-\tau)}{m}
\end{equation}

\section{Experimentation}

In order to test the effectiveness of the adaptive learning rate for mean absolute error and check loss, the scheme is tested against the commonly used datasets that consist of regression based modelling\footnote{All datasets used are open-source and obtained from scikit-learn and University of California, Irvine Machine Learning Repository(https://archive.ics.uci.edu/ml/index.php) 
}. 
\begin{enumerate}
    \item California Housing Dataset- This dataset consists of 20 443 samples and 9 features that can be used to predict the mean housing price in a particular locality. The features include location of the house, number of rooms within that particular locality, age of houses, among many others. 
    \item Boston Housing Dataset- The dataset consists of 13 features, which are used to predict the median house value in a particular locality. Some of the features include average number of rooms per house and crime rate.  
    \item Energy Efficiency Dataset- The dataset consists of 8 features, which are used to predict heating and cooling load requirements of buildings. The dataset is used to perform multivariate regression.  
\end{enumerate}{}

In order to quantitatively compare the effectiveness of the constant learning rate and adaptive learning rate schemes, two methods of comparison are used 
\begin{enumerate}
    \item \textbf{Number of epochs}- A threshold value $T_{L}$ is chosen and the number of epochs required for the constant and adaptive learning rate schemes to reach the threshold is measured. This is to measure which method leads to \textit{faster convergence}
    \item \textbf{Performance}- The model is trained and the loss value after a fixed number of epochs is compared for the constant and adaptive learning rate schemes. This is to determine which method results in a model with \textit{higher accuracy} in its predictions. 
\end{enumerate}{}

It is important to note that during the comparison, the \textit{same initial weights are used for both the schemes}. Also, a constant learning rate of 0.1 was used.

\subsection{Implementation details}
The entire framework is implemented in Keras, which uses Tensorflow as its backend. Keras allows for easy and fast prototyping.  The flexibility provided by Keras to add dense layers and dropout layers with minimum ease made it an obvious choice to experiment with the regression datasets. Depending on the dataset, the underlying architecture such as the number of dense layers, number of neurons in each layer and dropout layers is chosen accordingly. Learning rate is calculated at the beginning of each epoch based on the mathematical expression derived. Learning rate is easily integrated in the pipeline using the callbacks functionality in Keras.

\subsection{Choosing the threshold}
A regression model is constructed using Ordinary Least Squares in order to obtain the threshold. The intuition behind using regression models is to obtain estimates of convergence points that can be obtained using a neural network.

\section{Results}
\subsection{Mean Absolute Error}
Table \ref{results_summary_epochs} summarizes the results obtained for the three datasets
\begin{table}[h!]
    \caption{Mean Absolute Error: Number of epochs}
    \label{results_summary_epochs}
\resizebox{\columnwidth}{!}{%
    \begin{tabular}{|p{3.0cm}|p{1.5cm}|p{1.5cm}|p{1.5cm}|}
        \hline
        \textbf{Dataset} & \textbf{Threshold} & \textbf{Epochs \newline (constant)} & \textbf{Epochs \newline (adaptive)} \\
        \hline 
        California Housing & 0.371  & 1409  & 114 \\
        \hline
        Boston Housing  & 0.257 & 946 & 346 \\
        \hline
        Energy Efficiency & 0.229 & 1950 & 85 \\
        \hline
    \end{tabular}
}

\end{table}{}

\textit{Note: For Boston Housing and Energy Efficiency datasets, alternate methods of threshold calculations are used as stated below}

\begin{enumerate}
        \item Find the minimum loss value from the constant learning rate scheme 
        \item Set the threshold as the above mentioned loss value 
        \label{threshold_calculation}
        \captionof{threshold_calculation}{Alternative method for threshold calculation}
\end{enumerate}{}

Tables \ref{results_summary_loss} and \ref{results_summary_val_loss} compare the loss values after a fixed number of iterations of training between the constant and adaptive learning rate schemes. It is important to note that the adaptive learning rate automatically decreases over time for all the data sets, as proved mathematically in section V. 
\begin{table}[h!]
    \caption{Mean Absolute Error: Loss value}
    \label{results_summary_loss}
\resizebox{\columnwidth}{!}{%
    \begin{tabular}{|p{2.6cm}|p{1cm}|p{2cm}|p{2cm}|}
        \hline
        \textbf{Dataset} & \textbf{Epochs} & \textbf{Loss(constant)} & \textbf{Loss(adaptive)} \\
        \hline 
        California Housing & 2500  & 0.3630 $\pm$ 0.0049  & 0.3474 $\pm$ 0.0031 \\
        \hline
        Boston Housing & 1000 & 0.2621 $\pm$ 0.0015 & 0.2480 $\pm$ 0.0021 \\
        \hline
        Energy Efficiency & 2000 & 0.2282 $\pm$ 0.0020   & 0.1631 $\pm$ 0.0042  \\
        \hline
    \end{tabular}
}

\end{table}{}

\begin{table}[h!]
    \caption{Mean Absolute Error: Validation Loss value}
    \label{results_summary_val_loss}
\resizebox{\columnwidth}{!}{%
    \begin{tabular}{|p{2.6cm}|p{1cm}|p{2cm}|p{2cm}|}
        \hline
        \textbf{Dataset} & \textbf{Epochs} & \textbf{Validation Loss(constant)} & \textbf{Validation Loss(adaptive)} \\[5pt]
        \hline 
        California Housing & 2500  & 0.3832 $\pm$ 0.0053  & 0.3686 $\pm$ 0.0038 \\ [10pt]
        \hline
        Boston Housing  & 1000 & 0.3118 $\pm$ 0.0096 & 0.3134 $\pm$ 0.0132 \\
        \hline
        Energy Efficiency & 2000 & 0.2537 $\pm$ 0.0037  & 0.1785 $\pm$ 0.0032  \\
        \hline
    \end{tabular}
}

\end{table}{}

\subsubsection{California Housing Dataset}

\begin{center}
    \begin{figure}[h]
        \centering
        \includegraphics[scale=0.48]{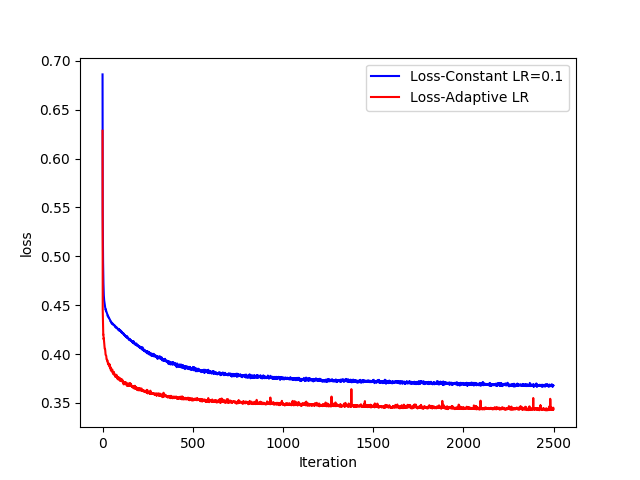}
        \caption{California Housing: Training loss over time}
        \label{loss_compare_CaliforniaHousing}
\end{figure}
\end{center}

\begin{center}
\begin{figure}[h]
    \centering
    \includegraphics[scale=0.48]{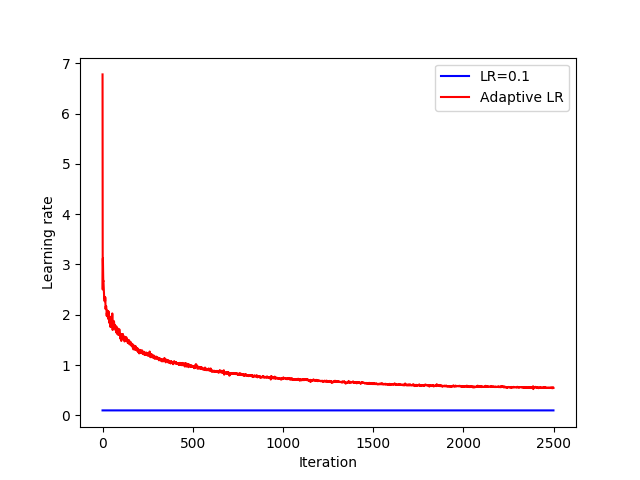}
    \caption{California Housing: Learning rate over time} 
    \label{lr_compare_CaliforniaHousing}
\end{figure}    
\end{center}

The architecture used for the California Housing dataset is described Table \ref{hyperparameters_CaliforniaHousing}. As noted in Figure \ref{lr_compare_CaliforniaHousing}, the learning rate starts at a large value of 6.78 and decreases exponentially before saturating at a value of 0.55. As depicted in Figure \ref{loss_compare_CaliforniaHousing}, the adaptive learning rate policy converges 10 times faster than the constant learning rate based model.  


\begin{table}[t]
    \caption{California Housing: Configuration of hyperparameters}
    \label{hyperparameters_CaliforniaHousing}
\resizebox{\columnwidth}{!}{%
    \begin{tabular}{|c|c|}
        \hline
        \textbf{Hyperparameter} & \textbf{Value} \\[5pt]
        \hline 
        Feature Scaling technique & Standardization \\ 
        \hline
        Batch size & 256 \\
        \hline
        Activation function & ReLU activation in the hidden layers\\ & SoftSign activation in the last layer \\
        \hline
        Optimization algorithm & Mini-batch Gradient Descent\\
        \hline        
        Number of hidden layers & 2\\
        \hline
        Number of hidden neurons & [20,15] \\
        \hline
        Number of output units & 1 \\
        \hline
    \end{tabular}
}

\end{table}{}

{}

{}

\subsubsection{Energy Efficiency dataset}
\begin{center}
    \begin{figure}[t]
        \centering
        \includegraphics[scale=0.45]{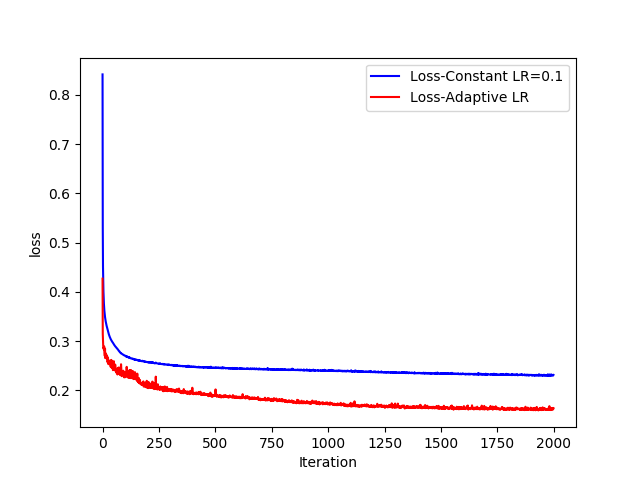}
        \caption{Energy Efficiency: Training loss over time}
        \label{loss_compare_EnergyEfficiency}
\end{figure}
\end{center}

Since the model predicts two output variables,the Lipschitz constant derived for multivariate regression in Section \RomanNumeralCaps{5B} is used to compute the adaptive learning rate. The architecture of the neural network consists of a single hidden layer with 50 neurons. The model is trained using Gradient Descent with a batch size of 64. The remaining hyperparameters are the same as the ones described in Table \ref{hyperparameters_CaliforniaHousing}. Figure \ref{loss_compare_EnergyEfficiency} shows that the loss due to the adaptive learning rate decreases faster than that due to the constant learning rate. Moreover, although the learning rate starts at a high value 5.87, it decreases rapidly and eventually leads to faster convergence.





{}


\subsubsection{Boston Housing Dataset}
The dataset is used to perform regression in order to predict housing prices in Boston. 
When implemented using a neural network, the architecture is described in Table \ref{hyperparameters_BostonHousing}. Although the adaptive learning rate scheme converges twice as fast as the constant learning rate policy, the performance improvement observed is relatively smaller compared to the improvement in the other datasets. 

In addition, the computational training time per epoch is larger for the adaptive learning rate based models due to the calculation of the adaptive learning rate. However, due to its faster convergence, LALR based models exhibit a smaller overall training time. For example, LALR based models trained on the Boston Housing dataset take up to 1.7 times longer to train per epoch compared to constant learning rate based models, but depict a reduction in the overall training time by upto 2 times. Similarly, LALR based models trained on the Energy Efficiency dataset and the California Housing dataset depict a reduction in training time by 6 times and 8 times respectively.   

\begin{table}[b]
    \small
    \caption{Boston Housing: Configuration of hyperparameters}
    \label{hyperparameters_BostonHousing}
\resizebox{\columnwidth}{1.6cm}{%
    \begin{tabular}{|c|c|}
        \hline
        \textbf{Hyperparameter} & \textbf{Value} \\[5pt]
        \hline 
        Feature Scaling technique & Standardization \\ 
        \hline
        Batch size & 8 \\
        \hline
        Activation function & ReLU activation in the hidden layer \\
        & SoftSign activation in the last layer \\
        \hline
        Optimization algorithm & Mini-batch Gradient Descent\\
        \hline        
        Number of hidden layers & 1\\
        \hline
        Number of hidden neurons & [20]\\
        \hline
        Number of output units & 1\\
        \hline
    \end{tabular}
}

\end{table}{}

\begin{center}
    \begin{figure}[t]
        \centering
        \includegraphics[scale=0.52]{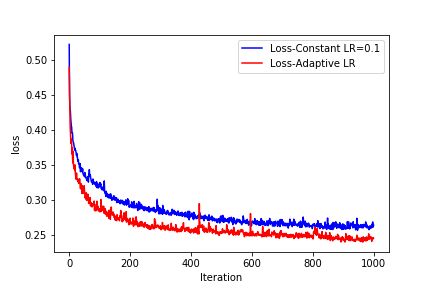}
        \caption{Boston Housing: Training loss over time}
        \label{loss_compare_BostonHousing}
\end{figure}
\end{center}{}

\begin{center}
    \begin{figure}[b]
        \centering
        \includegraphics[scale=0.52]{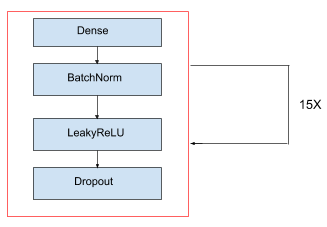}
        \caption{15 hidden layer network}
        \label{architecture_diagram_15}
\end{figure}
\end{center}

Although the network architectures considered are simplistic, experiments with deeper architectures depict no further improvement in the training and validation loss obtained. For example, when tested on a 15 hidden layer network, as shown in Figure \ref{architecture_diagram_15}, the adaptive learning rate based model exhibits a 
2.5x faster convergence than the constant learning rate based model for the California Housing Dataset. The 15-hidden layer network consists of 100 neurons in the first hidden layer and 50 hidden neurons in each subsequent hidden layer. The hidden activation unit used is LeakyReLU with a small slope of 0.3 on the negative side. In addition, a dropout of 10\% is used in each hidden unit. The remaining hyper-parameters are the same as shown in Table \ref{hyperparameters_CaliforniaHousing}. 

{}


\subsection{Quantile regression}
Firstly, quantile regression is implemented using neural networks in Keras and it is trained using a constant learning rate. The neural network consists of a single perceptron with zero hidden layers and a linear activation unit in its output layer. The network is trained using mini-batch gradient descent with a batch size of 64. The performance of the model is then compared against a baseline implementation in \texttt{statsmodel}\cite{seabold2010statsmodels}. 
Table \ref{performance_statsmodel} provides a summary of the results. The results indicate that the performance of neural network based learning is comparable to statistical based learning. 

\begin{table}[t]
    \caption{Comparison of quantile regression with a baseline implementation(AIC: Akaike Information Criteria)}
    \label{performance_statsmodel}
\resizebox{\columnwidth}{!}{%
    \begin{tabular}{|p{1.1cm}|p{1.1cm}|p{1.1cm}|p{1.1cm}|p{1.1cm}|p{1.1cm}|p{1.1cm}|}
        \hline
        \multirow{3}{10cm}{\textbf{Dataset}} & \multicolumn{6}{|c|}{\textbf{Quantiles}} \\
        \cline{2-7} & \multicolumn{2}{|c|}{\textbf{5th}} & \multicolumn{2}{|c|}{\textbf{50th}} & \multicolumn{2}{|c|}{\textbf{95th}} \\
        \cline{2-7} & {\textbf{AIC}}  & {\textbf{AIC baseline}} & {\textbf{AIC}}  & {\textbf{AIC baseline}} & {\textbf{AIC}}  & {\textbf{AIC baseline}} \\
        \hline 
        California \newline Housing & 1278.98  & 1279.05 & 6180.87 & 6178.84 & 2298.53 & 2296.95 \\
        \hline
        Boston \newline Housing & 53.9094 & 51.8603 & 146.3936 & 145.3413 & 73.1767 & 73.2916  \\
        \hline
    \end{tabular}
}
\end{table}{}


Secondly, the efficacy of neural network based quantile regression is analyzed using a synthetically generated dataset. The dataset consists of a single feature and a single output and it is depicted in the following equation: $y=f(x) + \epsilon$  where $\epsilon \sim $ N(0,$\sigma(x)^{2}$); \quad $\sigma(x)$ = $0.1$ exp(1-x)  \cite{takeuchi2006nonparametric}.
Various quantiles are chosen and individual neural networks are trained with the Check loss function for each quantile. The neural network consists of 2 hidden layers, with 10 and 5 neurons respectively. The hidden activation units employ the softplus activation function  and a linear activation unit in the output layer. The network is trained using mini-batch gradient descent for 3000 iterations with a batch size of 64. The results are summarized in Table \ref{simulated_dataset}. Hence, neural networks are effective in modelling heteroscedastic data for quantile regression. 

\begin{table}[t]
    \caption{Comparison of theoretical and predicted quantiles}
    \label{simulated_dataset}
\resizebox{\columnwidth}{!}{%
    \begin{tabular}{|P{3.5cm}|P{3.5cm}|}
        \hline
        \textbf{Theoretical quantile} & \textbf{Predicted quantile} \\
        \hline
        0.05 & 0.07 \\
        \hline
        0.30 & 0.296 \\
        \hline
        0.50 & 0.457 \\
        \hline
        0.70 & 0.732 \\
        \hline
        0.95 & 0.959 \\
        \hline
    \end{tabular}
}
\end{table}{}
Due to the effectiveness of neural network based quantile regression, an adaptive learning rate policy for quantile loss is tested. To elaborate, the adaptive learning rate scheme is tested against a constant learning rate of 0.1 for the datasets mentioned in Section \RomanNumeralCaps{6}. Experiments are run independently for each chosen value of quantile and the results are summarized in Tables \ref{results_summary_epochs_quantile} \ref{results_summary_loss_quantile}, \ref{results_summary_val_loss_quantile}. The threshold is calculated according to Heuristic 1 for all the datasets. Furthermore, the architecture and hyperparameters used for each of the datasets are identical to the ones used with Mean Absolute Error, as mentioned in Section \RomanNumeralCaps{7}A, unless specified otherwise.

\begin{table}[t]
    \caption{Quantile loss: Comparison of iterations}
    \label{results_summary_epochs_quantile}
\resizebox{\columnwidth}{!}{%
    \begin{tabular}{|p{2cm}|p{1cm}|p{1cm}|p{1cm}|p{1cm}|p{1cm}|p{1cm}|}
        \hline
        \multirow{2}{2cm}{\textbf{Dataset}} & \multicolumn{2}{|c|}{\textbf{Threshold}} & \multicolumn{2}{|c|}{\textbf{Epochs(constant LR)}} & \multicolumn{2}{|c|}{\textbf{Epochs(adaptive LR)}} \\

        \cline{2-7} & {\textbf{5th}}  & {\textbf{95th}} & {\textbf{5th}}  & {\textbf{95th}} & {\textbf{5th}}  & {\textbf{95th}} \\
        \hline 
        California \newline Housing & 0.059 & 0.1659 & 995 & 1000 & 138 & 92 \\
        \hline
        Boston Housing  & 0.0755 & 0.1605 & 1000 & 1000 & 47 & 42  \\
        \hline
        Energy Efficiency  & 0.0406 & 0.1110 & 1977 & 1996 & 144 & 184 \\
        \hline
    \end{tabular}
}

\end{table}{}

\begin{table}[t]
    \caption{Quantile loss: Loss value for various quantiles}
    \label{results_summary_loss_quantile}
\resizebox{\columnwidth}{!}{%
    \begin{tabular}{|p{3cm}|p{1cm}|p{1cm}|p{1cm}|p{1cm}|p{1cm}|}
        \hline
        \multirow{2}{2cm}{\textbf{Dataset}} & \multirow{2}{2cm}{\textbf{Epochs}} & \multicolumn{2}{|c|}{\textbf{Loss(constant LR)}} & \multicolumn{2}{|c|}{\textbf{Loss(adaptive LR)}} \\
        & &  {\textbf{5th}} & {\textbf{95th}} & {\textbf{5th}} & {\textbf{95th}}  \\
        \hline
        California  Housing & 1000  &  0.0582 $\pm$ \newline 0.0006 & 0.1659 $\pm$ \newline 0.0004  & 0.0551  \newline $\pm$ \newline 0.0008 & 0.1599 $\pm$ \newline 0.0004  \\
        \hline
        Boston Housing & 1000 & 0.0749 $\pm$ \newline 0.0008 & 0.1597 $\pm$ \newline 0.0007 & 0.0656 $\pm$ \newline 0.0008 & 0.1439 $\pm$ \newline 0.0015 \\
        \hline
        Energy Efficiency & 2000  & 0.0407 $\pm$ \newline 0.0002 & 0.1107 $\pm$ \newline 0.0002 & 0.0357 $\pm$ \newline 0.0004 & 0.1062 $\pm$ \newline 0.0002 \\
        \hline
    \end{tabular}
}

\end{table}{}

\begin{table}[h!]
    \caption{Quantile validation loss: Validation Loss value for various quantiles}
    \label{results_summary_val_loss_quantile}
\resizebox{\columnwidth}{!}{%
    \begin{tabular}{|p{2cm}|p{1cm}|p{1cm}|p{1cm}|p{1cm}|p{1cm}|}
        \hline
        \multirow{2}{2cm}{\textbf{Dataset}} & \multirow{2}{2cm}{\textbf{Epochs}} & \multicolumn{2}{|c|}{\textbf{Loss(constant LR)}} & \multicolumn{2}{|c|}{\textbf{Loss(adaptive LR)}} \\
        & &  {\textbf{5th}} & {\textbf{95th}} & {\textbf{5th}} & {\textbf{95th}}  \\
        \hline
        California \newline Housing & 1000  & 0.0616  $\pm$ \newline 0.0007 & 0.1654 $\pm$ \newline 0.0004  & 0.0595  $\pm$ \newline 0.0009 & 0.1621 $\pm$ \newline 0.0005  \\
        \hline
        Boston Housing & 1000 & 0.0744 $\pm$ \newline 0.0009 & 0.1401 $\pm$ \newline 0.0019 & 0.0695 $\pm$ \newline 0.0030 & 0.1377 $\pm$ \newline 0.0015 \\
        \hline
        Energy Efficiency$^{\mathrm{a}}$ & 2000 & 0.0475 $\pm$ \newline 0.0005 & 0.1130 $\pm$ \newline 0.0006 & 0.0424 $\pm$ \newline 0.0009 & 0.1081 $\pm$ \newline 0.0003 \\
        \hline
        \multicolumn{6}{c}{$^{\mathrm{a}}$ Multiple Regression is used to predict a single output variable}
\end{tabular}
}

\end{table}{}

\begin{figure*}[h!]
    \begin{subfigure}[t]{0.5\textwidth}
        \includegraphics[scale=0.40]{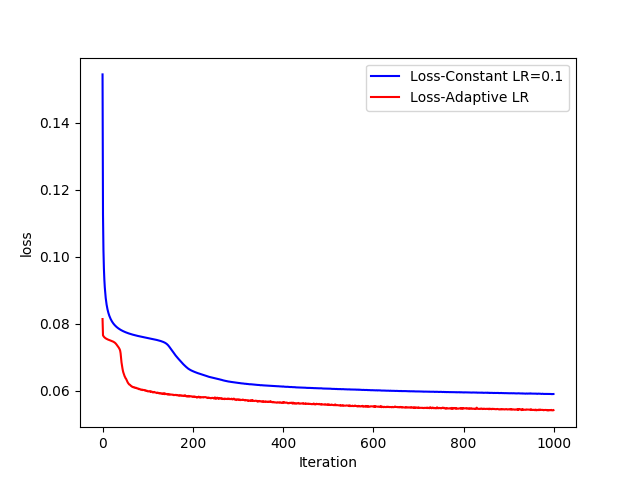}
        \caption{$5^{th}$ quantile for California Housing: Training loss}
        \label{loss_compare_CaliforniaHousing_5}
    \end{subfigure}
    \hfill
    \begin{subfigure}[t]{0.5\textwidth}
        \includegraphics[scale=0.40]{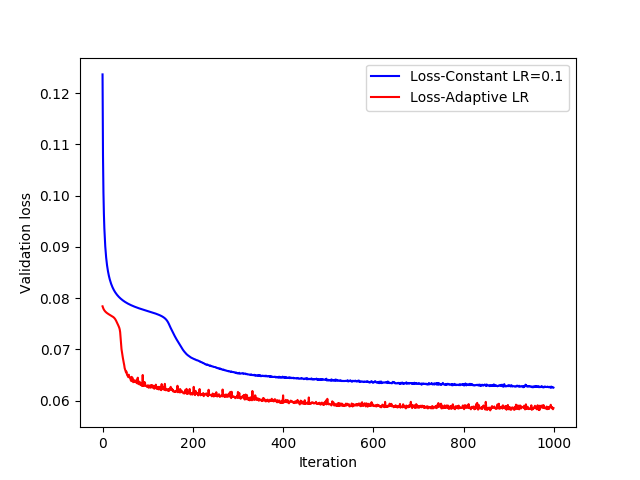}
        \caption{$5^{th}$ quantile for California Housing: Validation loss}
        \label{val_loss_compare_CaliforniaHousing_5}
    \end{subfigure}
    \hfill
    \begin{subfigure}[t]{0.5\textwidth}
        \includegraphics[scale=0.40]{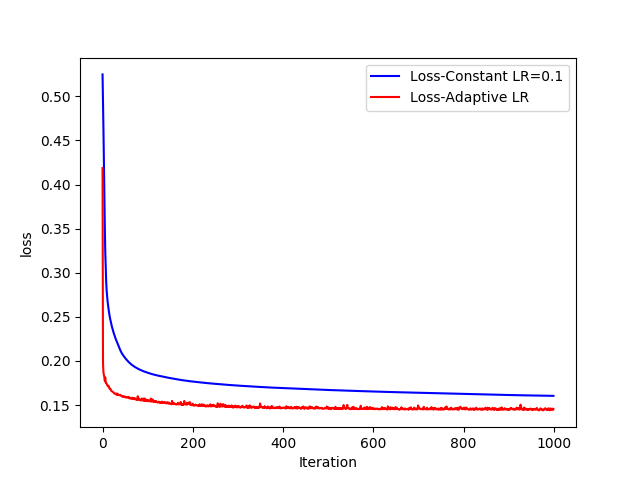}
        \caption{$95^{th}$ quantile for Boston Housing: Training loss}
        \label{loss_compare_BostonHousing_95}
    \end{subfigure}
    \hfill
    \begin{subfigure}[t]{0.5\textwidth}
        \includegraphics[scale=0.40]{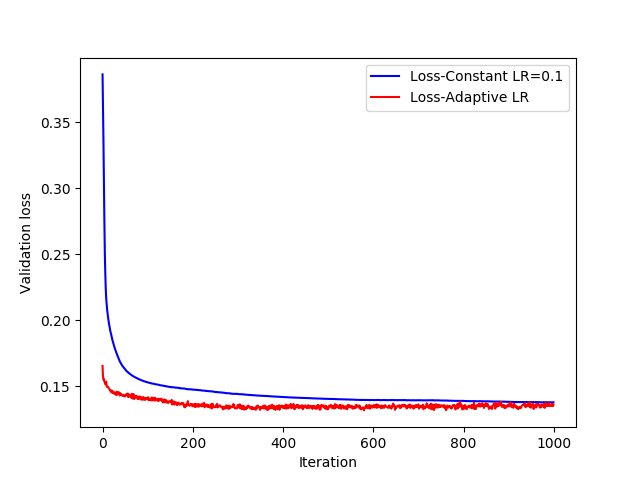}
        \caption{$95^{th}$ quantile for Boston Housing: Validation loss}
        \label{val_loss_compare_BostonHousing_95}
    \end{subfigure}
    \caption{Training and validation loss over time}
    \label{loss_compare_BostonHousing_EnergyEffiency_quantile}
\end{figure*}

The adaptive learning rate for Check loss converges up to 13 times faster than the constant learning rate for the $5^{th}$ quantile and up to 10 times faster for the $95^{th}$ quantile. 
Figures \ref{loss_compare_CaliforniaHousing_5} and \ref{val_loss_compare_CaliforniaHousing_5} depict the training and validation loss for the California Housing dataset for the  $5^{th}$ quantile. It is important to note that the model is able to achieve faster convergence without overfitting. Furthermore, the models for Energy Efficiency dataset also depict a similar trend with a speed up in convergence by 11 times. The architecture proposed in Table \ref{hyperparameters_BostonHousing} for the Boston Housing dataset resulted in overfitting of the model. Consequently, the architecture is slightly modified to a single hidden layer network with 15 neurons. Moreover, the model is trained with a batch size of 256 for 1000 epochs. The remaining hyperparameters remain unchanged. Figures \ref{loss_compare_BostonHousing_95} and \ref{val_loss_compare_BostonHousing_95} depict a faster convergence by nearly 20 times, a significant increase in the speed of convergence compared to the other two datasets. In addition, the adaptive learning rate for the Boston Housing dataset starts at a large value of 6.43 and saturates at a value of 2.58. In contrast, the adaptive learning rate for the Energy Efficiency dataset starts at a relatively lower value of 3.23 and saturates at a value of 0.76. 

Furthermore, as mentioned in Section \RomanNumeralCaps{7}A, although additional computation is performed during every epoch in order to calculate the adaptive learning rate, the training time for LALR based models is lower than constant learning rate based models. For instance, LALR based models trained on the Energy Efficiency dataset depict an increase in training time per epoch by an average of 5 times, but exhibit an overall decrease in training time by 8 times. Similarly, for the Boston Housing dataset, LALR based models depict a decrease in the overall training time by up to 5 times although the computational time per epoch is 7 times higher.

\section{Conclusion and Future Work}
In summary, the paper proposes an adaptive learning rate scheme for regression based problems that utilize mean absolute error and check loss as loss functions. A theoretical framework for Lipschitz learning rate is derived for MAE and Check loss and its effectiveness is evaluated against commonly used regression based datasets. 

It is found that the adaptive learning rate policy  performs better than the constant learning rate policy by a significant amount. For mean absolute error, the adaptive learning rate increases the speed of convergence by 5 to 20 times. Similarly, the adaptive learning rate for Check loss also achieves faster convergence by nearly 20 times. It is important to note that even though the adaptive learning rate begins at a large value, faster convergence and better performance is achieved with Stochastic Gradient Descent, along with a natural decay in the learning rate. Hence, uncertainty in the prediction estimates of deep neural networks can be obtained with a smaller training time, thus increasing the reliability of the predictions.

An area of future work is to extend the theoretical framework for various other optimization algorithms such as momentum based gradient descent and Adam and evaluate its effectiveness. Furthermore, we also wish to explore an adaptive learning rate for Check loss that handles quantile crossing. Another area of future work is to analyze the relationship between Lipschitz Adaptive learning rate and activation units used in neural networks, primarily in the context of exploding gradients.

\section*{Acknowledgement}
The authors would like to thank the Science and Engineering Research Board (SERB)-DST, Government of of India for supporting this research (File SERB-EMR/ 2016/005687). 

\bibliography{citations,./IEEEabrv}



\end{document}